\documentclass{article}

\usepackage{times}
\usepackage{graphicx} % more modern
\usepackage{subfigure} 

\usepackage{natbib}

\usepackage{algorithm}
\usepackage{algorithmic}

\usepackage{hyperref}

\usepackage[accepted]{icml2017}

\icmltitlerunning{Statistical Inference for Incomplete Ranking Data}

\usepackage{mathtools}
\usepackage{bm}
\usepackage{microtype}

\usepackage{amsmath}
\usepackage{amssymb} 
\usepackage{amsthm}
\usepackage{graphicx}

\def\S{\mathbb{S}}

\DeclareMathOperator*{\argmax}{\arg \max}
\DeclareMathOperator*{\argmin}{\arg \min}
\newcommand{\given}{\, | \,}
\newcommand{\sign}{\operatorname{sign}}
\newcommand{\Prob}{P}
\newcommand{\prob}{\boldsymbol{p}}
\newcommand{\p}{\boldsymbol{p}}
\renewcommand{\vec}[1]{\boldsymbol{#1}}

\newcommand{\fromto}{\longrightarrow}
\newcommand{\argsort}{\operatorname*{arg\,sort}}

\newcommand{\pl}{\boldsymbol{pl}}

\begin{document} 
	
	\twocolumn[
	\icmltitle{Statistical Inference for Incomplete Ranking Data:\\ The Case of Rank-Dependent Coarsening}
	
	
	
	% you can specify symbols, otherwise they are numbered in order
	\icmlsetsymbol{equal}{*}
	
	\begin{icmlauthorlist}
		\icmlauthor{Mohsen Ahmadi Fahandar}{to}
		\icmlauthor{Eyke H\"ullermeier}{to}
		\icmlauthor{In\'es Couso}{goo}
	\end{icmlauthorlist}
	
	\icmlaffiliation{to}{Paderborn University, Germany}
	\icmlaffiliation{goo}{University of Oviedo, Spain}
	
	\icmlcorrespondingauthor{Eyke H\"ullermeier}{eyke@upb.de}

	\icmlkeywords{preference learning, ranking, Plackett-Luce, coarse data, consistency}
	
	\vskip 0.3in
	]
	
	
	
	\printAffiliationsAndNotice{}  % leave blank if no need to mention equal contribution
	%\footnotetext{hi}

\begin{abstract}

	We consider the problem of statistical inference for ranking data, specifically rank aggregation, under the assumption that samples are incomplete in the sense of not comprising all choice alternatives. In contrast to most existing methods, we explicitly model the process of turning a full ranking into an incomplete one, which we call the \emph{coarsening process}. To this end, we propose the concept of \emph{rank-dependent} coarsening, which assumes that incomplete rankings are produced by projecting a full ranking to a random subset of ranks. For a concrete instantiation of our model, in which full rankings are drawn from a Plackett-Luce distribution and observations take the form of pairwise preferences, we study the performance of various rank aggregation methods. In addition to predictive accuracy in the finite sample setting, we address the theoretical question of consistency, by which we mean the ability to recover a target ranking when the sample size goes to infinity, despite a potential bias in the observations caused by the (unknown) coarsening. 
\end{abstract}

\vspace*{-5mm}

\section{Introduction}
\label{intro}

The analysis of rank data has a long tradition in statistics, and corresponding methods have been used in various fields of application, such as psychology and the social sciences \cite{mard_aa}. More recently, applications in information retrieval and machine learning have caused a renewed interest in the analysis of rankings and topics such as ``learning-to-rank'' \cite{liu_lt} and preference learning \cite{mpub218}.

In most applications, the rankings observed are \emph{incomplete} or \emph{partial} in the sense of including only a subset of the underlying choice alternatives (subsequently referred to as ``items''), whereas no preferences are revealed about the remaining ones---pairwise comparisons can be seen as an important special case.
Somewhat surprisingly, most methods for learning from ranking data, including methods for rank aggregation, simply ignore the process of turning a full ranking into an incomplete one. Or, they implicitly assume that the process is unimportant from a statistical point of view, because the subset of items observed is independent of the underlying ranking. 

Obviously, this assumption is often not valid, as shown by practically relevant examples such as top-$k$ observations. Motivated by examples of this kind, we propose the concept of \emph{rank-dependent coarsening}, which assumes that incomplete rankings are produced by projecting a full ranking to a random subset of ranks. The notion of ``coarsening'' is meant to indicate that an incomplete ranking can be associated with a set of complete rankings, namely the set of its consistent extensions---set-valued data of that kind is also called ``coarse data'' in statistics \cite{coarse91, coarse97}. The idea of coarsening is similar to the interpretation of partial rankings as ``censored data'' \cite{leba_nm08}. The assumption of rank-dependent coarsening can be seen as orthogonal to standard marginalization, which acts on items instead of ranks \cite{rajk_as14,sibo_mb15}.

In addition to introducing a general statistical framework for analyzing incomplete ranking data (Section~3), 
we outline several problems and learning tasks to be addressed in this framework. Learning of the entire model will normally not be feasible, even under restrictive assumptions on the coarsening. A specifically interesting question, therefore, is to what extent and in what sense successful learning is possible for methods that are agnostic of the coarsening process. We investigate this question, both practically (Section~6) and theoretically (Section~7), for several ranking methods (Section~5) and a concrete instantiation of our framework, in which full rankings are drawn from a Plackett-Luce distribution and observations take the form of pairwise preferences (Section~4). In particular, we are interested in the property of consistency, by which we mean the ability to recover a target ranking when the sample size goes to infinity, despite a potential bias in the observations caused by the coarsening.

\section{Preliminaries and Notation}

Let $\S_K$ denote the collection of rankings (permutations) over a set $U=\{a_1, \ldots, a_K\}$ of  $K$ items $a_k$, $k \in [K] = \{1, \ldots, K\}$.  We denote by $\pi: [K] \fromto [K]$ a complete ranking (a generic element of $\S_K$), where $\pi(k)$ denotes the position of the $k^{th}$ item $a_k$ in the ranking, and by $\pi^{-1}$ the ordering associated with a ranking, i.e., $\pi^{-1}(j)$ is the index of the item on position $j$. We write rankings in brackets and orderings in parentheses; for example, $\pi=[2,4,3,1,5]$ and $\pi^{-1} = (4,1,3,2,5)$ both denote the ranking $a_4  \succ  a_1  \succ  a_3  \succ  a_2  \succ  a_5$. 

For a possibly incomplete ranking, which includes only some of the items, we use the symbol $\tau$ (instead of $\pi$). If the $k^{th}$ item does not occur in a ranking, then $\tau(k)=0$ by definition; otherwise, $\tau(k)$ is the rank of the $k^{th}$ item. In the corresponding ordering, the missing items do simply not occur. For example, the ranking $a_4 \succ a_1 \succ a_2$ would be encoded as $\tau=[2,3,0,1]$ and $\tau^{-1} = (4,1,2)$, respectively. We let $I(\tau) = \{ k \, : \, \tau(k) > 0 \} \subset [K]$ and denote the set of all rankings (complete or incomplete) by $\overline{\S}_K$.

An incomplete ranking $\tau$ can be associated with its set of linear extensions $E(\tau) \subset \S_K$, where $\pi \in E(\tau)$ if $\pi$ is consistent with the order of items in $I(\tau)$, i.e., $(\tau(i)-\tau(j))(\pi(i)- \pi(j)) \geq 0$  for all  $i , j \in I(\tau)$. 
An important special case is an incomplete ranking $\tau = \tau_{i,j}=(i,j)$ in the form of a pairwise comparison $a_i\succ a_j$ (i.e., $\tau(i)=1$, $\tau(j)=2$, $\tau(k) = 0$ otherwise), which is associated with the set of extensions 
$$
E(\tau) = E(a_i \succ a_j) =  \{\pi\in\S_K\,:\, \pi(i)<\pi(j)\} \enspace .
$$
Modeling an incomplete observation $\tau$ by the set of linear extensions $E(\tau)$ reflects the idea that $\tau$ has been produced from an underlying complete ranking $\pi$ by some ``coarsening'' or ``imprecisiation'' process, which essentially consists of omitting some of the items from the ranking. $E(\tau)$ then corresponds to the set of all consistent extension $\pi$ if nothing is known about the coarsening, except that it does not change the relative order of any items.

\section{General Setting and Problems}

The type of data we assume as observations is incomplete rankings $\tau \in \overline{\S}_K$. Statistical inference for this type of data requires a probabilistic model of the underlying data generating process, that is, a probability distribution on $\overline{\S}_K$.

\subsection{A Stochastic Model for Incomplete Rankings}

Recalling our idea of a coarsening process, it is natural to consider the data generating process as a two step procedure, in which a full ranking $\pi$ is generated first and turned into an incomplete ranking $\tau$ afterward. We model this assumption in terms of a distribution on $\overline{\S}_K \times \S_K$, which assigns a degree of probability to each pair $(\tau , \pi)$. More specifically, we assume a parameterized distribution of the following form:
\begin{equation}\label{eq:dgp}
\prob_{\theta , \lambda}(\tau, \pi ) = \prob_\theta(\pi ) \cdot \prob_\lambda(\tau \given \pi ) 
\end{equation}
Thus, while the generation of full rankings is determined by the distribution 
\begin{equation}\label{eq:pp}
\prob_\theta:\, \S_K \fromto [0,1] \enspace ,
\end{equation}
the coarsening process is specified by a family of conditional probability distributions
\begin{equation}\label{eq:cp}
\big\{ \prob_\lambda(\cdot \given \pi) \, : \, \pi \in \S_K , \, \lambda \in \Lambda \big\}  \enspace ,
\end{equation}
where $\lambda$ collects all parameters of these distributions; $\prob_{\theta , \lambda}(\tau, \pi )$ is the probability of producing the data $(\tau, \pi) \in \overline{\S}_K \times \S_K$. Note, however, that $\pi$ is actually not observed.

\subsubsection{Rank-Dependent Coarsening}

In its most general form, the coarsening process (\ref{eq:cp}) is extremely rich, even if being restricted by the consistency assumption $\prob_\lambda(\tau \given \pi) = 0$ for $\pi \not\in E(\tau)$.
In fact, since the number of probabilities to be specified is of the order $2^KK!$, inference about $\lambda$ will generally be difficult. Therefore, $\prob_\lambda$ certainly needs to be restricted by further assumptions. Apart from practical reasons, such assumptions are also indispensable for successful learning. Otherwise, observations could be arbitrarily biased in favor or disfavor of items, so that an estimation of the underlying (full) preferences, as reflected by $\prob_\theta$, becomes completely impossible. For example, the coarsening process may leave a ranking $\pi$ unchanged whenever item $a_1$ is on the last position, and remove $a_1$ from $\pi$ otherwise. Obviously, this item will then appear to have a very low preference. 

As shown by this example, the estimation of preferences will generally be impossible unless the coarsening is somehow more ``neutral''. The assumption we make here is a property we call \emph{rank-dependent} coarsening. A coarsening procedure is rank-dependent if the incompletion is only acting on \emph{ranks} (positions) but not on \emph{items}. That is, the procedure randomly selects a subset of ranks and removes the items on these ranks, independently of the items themselves. In other words, an incomplete observation $\tau$ is obtained by projecting a complete ranking $\pi$ on a random subset of positions $A \in 2^{[K]}$, i.e., the family (\ref{eq:cp}) of distributions $\prob_\lambda(\cdot \given \pi)$ is specified by a single measure on $2^{[K]}$. Or, stated more formally,
$$
\prob\big( \pi^{-1}(A) \given \pi^{-1} \big) = \prob\big( \sigma^{-1}(A) \given \sigma^{-1} \big)
$$
for all $\pi, \sigma \in \mathbb{S}^K$ and $A \subset [K]$, where $\pi^{-1}(A)$ denotes the projection of the ordering $\pi^{-1}$ to the positions in $A$. 

The assumption of rank-dependent coarsening can be seen as orthogonal to standard marginalization: while the latter projects a full ranking to a \emph{subset of items}, the former projects a ranking to a \emph{subset of positions}. The practically relevant case of top-$k$ observations is a special (degenerate) case of rank-dependent coarsening, in which
$$
\prob(A) = \left\{ \begin{array}{cl}
1 & \text{ if } A = \{1, \ldots , k \} \\
0 & \text{ otherwise} 
\end{array} \right.
$$
This model could be weakened in various ways. For example, instead of fixing the length of observed rankings to a constant, $k$ could be a random variable. Or, one could assume that positions are discarded with increasing probability, though independently of each other; thus, the probability to observe a subset of items on ranks $A \subseteq \{1, \ldots , K\}$ is given by
$$
\Prob(A) = \prod_{i \in A} \lambda_i \cdot \prod_{j \not\in A} (1-\lambda_j) \, .
$$
The coarsening is then defined by the $K$ parameters $\lambda_1 > \lambda_2 > \ldots > \lambda_K$.

\subsection{Learning Tasks}

Suppose a sample of (training) data $\mathcal{D} = \{ \tau_1, \ldots , \tau_N\}$ to be given. As for statistical inference about the process (\ref{eq:dgp}), several problems could be tackled.

\begin{itemize}
	\item
	The most obvious problem is to estimate the complete distribution on $\overline{\S}_K$, i.e., the parameters $\theta$ and $\lambda$. As already explained before, this will require specific assumptions about the coarsening. 
	
	\item
	A somewhat weaker goal is to estimate the ``precise part'', i.e., the parameter $\theta$. Indeed, in many cases, $\theta$ will be the relevant part of the model, as it specifies the preferences on items, whereas the coarsening is rather considered as a complication of the estimation. In this case, $\lambda$ is of interest only in so far as it helps to estimate $\theta$. Ideally, it would even be possible to estimate $\theta$ without any inference about $\lambda$, i.e., by simply ignoring the coarsening process.

	\item
	An even weaker goal is to estimate, not the parameter $\theta$ itself, but only an underlying ``ground truth'' ranking $\pi^*$ associated with $\theta$. Indeed, in the context of learning to rank, the ultimate goal is typically to predict a ranking, not necessarily a complete distribution. For example, the ranking $\pi^*$ could be the mode of the distribution $\prob_\theta$, or any other sort of representative statistics. This problem is especially relevant in practical applications such as rank aggregation, in which $\pi^*$ would play the role of a consensus ranking.
	
\end{itemize}

Here, we are mainly interested in the third problem, i.e., the estimation of a ground-truth ranking $\pi^*$. Moreover, due to reasons of efficiency, we are aiming for an estimation technique that circumvents direct inference about $\lambda$, while being robust in the sense of producing reasonably good results for a wide range of coarsening procedures.

\section{Specific Setting and Problems}\label{sec:specific}

The development and analysis of methods is only possible for concrete instantiations of the setting introduced in the previous section. An instantiation of that kind will be proposed in this section. The first part of our data generating process, $\prob_\theta$, will be modeled by the Plackett-Luce model \cite{pl75, luce59}. To make the second part, $\prob_\lambda$, manageable, we restrict observations to the practically relevant case of pairwise comparisons (i.e., incomplete rankings of length 2).

\subsection{The Plackett-Luce Model}

The Plackett-Luce (PL) model is parameterized by a vector $\theta = (\theta_1, \theta_2, \ldots , \theta_K) \in \Theta = \mathbb{R}_+^K$. Each $\theta_i$ can be interpreted as the weight or ``strength'' of the option $a_i$. The probability assigned by the PL model to a ranking represented by a permutation $\pi \in \S_K$ is
given by
\begin{equation}\label{eq:pl}
\pl_\theta(\pi) =  \prod_{i=1}^K \frac{\theta_{\pi^{-1}(i)}}{\theta_{\pi^{-1}(i)} + \theta_{\pi^{-1}(i+1)} + \ldots + \theta_{\pi^{-1}(K)}}
\end{equation}
Obviously, the PL model is invariant toward multiplication of $\theta$ with a constant $c >0$, i.e., $\pl_\theta(\pi) = \pl_{c \theta}(\pi)$ for all $\pi \in \S_K$ and $c > 0$. Consequently, $\theta$ can be normalized without loss of generality (and the number of degrees of freedom is only $K-1$ instead of $K$). Note that the most probable ranking, i.e., the mode of the PL distribution, is simply obtained by sorting the items in decreasing order of their weight:
\begin{equation}\label{eq:mpr}
\pi^* = \argmax_{\pi \in \S_K} \pl_\theta (\pi ) = \argsort_{k \in [K]} \{ \theta_1, \ldots , \theta_K \} \, .
\end{equation}
As a convenient property of PL, let us mention that it allows for an easy computation of marginals, because the marginal probability on a subset $U' = \{ a_{i_1}, \ldots , a_{i_J} \} \subset U$ of $J \leq K$ items is again a PL model parametrized by $(\theta_{i_1}, \ldots , \theta_{i_J})$. Thus, for every $\tau \in \overline{\S}_K$ with $I(\tau) = U'$, 
%\begin{equation}\label{eq:plm}
$$
\pl_\theta (\tau)  =  \prod_{j=1}^J \frac{\theta_{\tau^{-1}(j)}}{\theta_{\tau^{-1}(j)} + \theta_{\tau^{-1}(j+1)} + \ldots + \theta_{\tau^{-1}(J)}}
$$
%\end{equation}
In particular, this yields pairwise probabilities 
\begin{equation}\label{eq:pmp}
p_{i,j} = \pl_\theta(\tau_{i,j}) = \frac{\theta_i}{\theta_i + \theta_j} \, ,
\end{equation}
where $\tau_{i,j} = (i,j)$ represents the preference $a_i \succ a_j$.
This is the well-known Bradley-Terry-Luce model \cite{brad_tr52}, a model for the pairwise comparison of alternatives. Obviously, the larger $\theta_i$ in comparison to $\theta_j$, the higher the probability that $a_i$ is chosen. The PL model can be seen as an extension of this principle to more than two items: the larger the parameter $\theta_i$ in (\ref{eq:pl}) in comparison to the parameters $\theta_j$, $j \neq i$, the higher the probability that $a_i$ appears on a top rank.

\subsection{Pairwise Preferences}

If rank-dependent coarsening is restricted to the generation of pairwise comparisons, the entire distribution $\prob_\lambda$ is specified by the set of $K(K-1)/2$ probabilities
\begin{equation}\label{eq:pc}
\!
\Big\{ \lambda_{i,j}  \given 1 \leq i < j \leq K , \, \lambda_{i,j} \geq 0, \!\!\sum_{1 \leq i < j \leq K} \lambda_{i,j} = 1 \Big\} \, ,
\end{equation}
where $\lambda_{i,j}$ denotes the probability that the ranks $i$ and $j$ are selected.

The problem of ranking based on pairwise comparisons has been studied quite extensively in the literature, albeit without taking coarsening into account, i.e., without asking where the pairwise comparisons are coming from (or implicitly assuming they are generated as marginals). Yet, worth mentioning is a recent study on \emph{rank breaking} \cite{souf_cp14}, that is, of the estimation bias (for models such as PL) caused by replacing full rankings in the training data by the set of all pairwise comparisons---our study of the bias caused by coarsening is very much in the same spirit.

\subsection{The Data Generating Process}

Combining the PL model (\ref{eq:pl}) with the coarsening process (\ref{eq:pc}), we obtain a distribution $\boldsymbol{q}$ on $\overline{\S}_K$ such that 
\begin{equation}\label{eq:pcp}
\boldsymbol{q}(\tau_{i,j}) = 
q_{i,j} =  \sum_{\pi \in E(a_i\succ a_j)} \pl_\theta(\pi) \, \lambda_{\pi(i), \pi(j)}
\end{equation} 
for pairwise preferences $\tau_{i,j}=(i,j)$, and $\boldsymbol{q}(\tau)=0$ otherwise.
Clearly, the pairwise probabilities (\ref{eq:pcp}) will normally not agree with the pairwise marginals (\ref{eq:pmp}). Instead, they may provide a biased view of the pairwise preferences between items. Please note, however, that the marginals $p_{i,j}$ are not directly comparable with the $q_{i,j}$, because the latter is a distribution on incomplete rankings ($\sum_{i,j} q_{i,j} = 1$) whereas the former is a set of marginal distributions ($p_{i,j} + p_{j,i} = 1$). Instead, $p_{i,j}$ should be compared to $q_{i,j}'= q_{i,j}/(q_{i,j} + q_{j,i})$, which is the probability that, in the coarsened model, $a_i$ is observed as a winner, given it is paired with $a_j$. 

As an illustration, consider a concrete example with $K=3$, $\theta = (14, 5,  1)$, and degenerate coarsening distribution specified by $\lambda_{1,2}=1$ (top-2 selection). One easily derives the probabilities of pairwise marginals (\ref{eq:pmp}) and coarsened (top-2) observations (\ref{eq:pcp}) as follows:
\begin{center}
	\begin{tabular}{l|cccccc}
		$i,j$ & $1,2$ & $1,3$ & $2,3$ & $2,1$ & $3,1$ & $3,2$ \\
		\hline \\[-2mm]
		$p_{i,j}$  & $\frac{840}{1140}$ & $\frac{1064}{1140}$ & $\frac{950}{1140}$ & $\frac{300}{1140}$ & $\frac{76}{1140}$ & $\frac{190}{1140}$ \\[2mm]
		$q_{i,j}$ & $\frac{665}{1140}$ & $\frac{133}{1140}$ & $\frac{19}{1140}$ & $\frac{266}{1140}$ & $\frac{42}{1140}$ & $\frac{15}{1140}$ \\[2mm]
		$q_{i,j}'$ & $\frac{665}{931}$ & $\frac{133}{175}$ & $\frac{19}{34}$ & $\frac{266}{931}$ & $\frac{42}{175}$ & $\frac{15}{34}$
	\end{tabular}
\end{center} 
While the $p_{i,j}$ are completely coherent with a PL model (namely $\pl_\theta$ with $\theta = (14, 5,  1)$), the $q_{i,j}$ and $q_{i,j}'$ no longer are.

A special case where coarsening is guaranteed to not introduce any bias is the uniform distribution $\lambda \equiv 2/(K^2 - K)$. In this case, random projection to ranks effectively coincides with random selection to items.

\subsection{Problems}

Of course, in spite of the inconsistency of the pairwise observations in the above example, a PL model could still be estimated, for example using the maximum likelihood principle. The corresponding estimate $\hat{\theta}$ will also yield an estimate $\hat{\pi}$ of the target ranking $\pi^*$ (which is simply obtained by sorting the items $a_i$ in decreasing order of the estimated scores $\theta_i$). As already said, there is little hope that $\hat{\theta}$ could be an unbiased estimate of $\theta$; instead $\hat{\theta}$ will necessarily be biased. There is hope, however, to recover the target ranking $\pi^*$. Indeed, the ranking will be predicted correctly provided $\hat{\theta}$ is comonotonic with $\theta$, i.e., $(\hat{\theta}_i - \hat{\theta}_j)(\theta_i - \theta_j) > 0$ for all $i,j \in [K]$. Roughly speaking, a small bias in the estimate can be tolerated, as long as the order of the parameters is preserved. 

Obviously, these considerations are not restricted to the PL model. Instead, any method for aggregating pairwise comparisons into an overall ranking can be used to predict $\pi^*$. This leads us to the following questions:
\begin{itemize}
	\item Practical performance: What is the performance of a rank aggregation method in the finite sample setting, i.e., how close is the prediction $\hat{\pi}$ to the ground truth $\pi^*$? How is the performance influenced by the coarsening process?
	
	\item Consistency: Is a method consistent in the sense that $\pi^*$ is recovered (with high probability) with an increasing sample size $N \rightarrow \infty$, either under specific assumptions on the coarsening process, or perhaps even regardless of the coarsening (i.e., only assuming the property of rank-dependence)?
\end{itemize}  

\section{Rank Aggregation Methods}

In this section, we discuss different rank aggregation methods that operate on pairwise data, %the performance of which has been investigated within our proposed setting. Algorithms are 
categorized according to the some basic principles. 
%under which they predict the overall ranking $\hat{\pi}$. It also allows us to evaluate different strategies for rank aggregation from pairwise data throughly and see how they work in case of our devised scenario. 
To this end, let us define the comparison matrix $C$ with entries $c_{i,j}$, where $c_{i,j}$ denotes the number of wins of $a_i$ over $a_j$, i.e., the number of times the preference $a_{i} \succ a_{j}$ is observed. Correspondingly, we define the probability matrix $\hat{P}$ with entries $\hat{p}_{i,j} = \frac{c_{i,j}}{ c_{i,j} + c_{j,i} }$, which can be seen as estimates of the winning probabilities $p_{i,j}$ (often also interpreted as weighted preferences). 
%\begin{equation*}
%\hat{P}_{i,j} =
%\begin{dcases*}
%\frac{c_{i,j}}{ c_{i,j} + c_{j,i} }  & if $i \ne j$,\\
%0 & if $i=j$.
%\end{dcases*}
%\end{equation*}
All rank aggregation methods produce rankings $\hat{\pi}$ based on either matrix $C$ or $\hat{P}$.

\subsection{Statistical Estimation}

\subsubsection{Bradley-Terry-Luce Model (BTL)}

The Bradley-Terry-Luce model is well-known in the literature on discrete choice \cite{brad_tr52}. It starts from the parametric model (\ref{eq:pmp}) of pairwise comparisons, i.e., the marginals of the PL model, and estimates the parameters by likelihood maximization:
$$
\hat{\theta} \in \arg \max_{\theta \in \mathbb{R}^{K} } \prod_{1 \leq i \neq j \leq K}  \left( \dfrac{\theta_{i}}{\theta_{i} + \theta_{j}} \right)^{c_{i,j}}
$$
The predicted ranking is obtained by sorting items according to their estimated strengths: $\hat{\pi} = \argsort (\hat{\theta})$.

As already explained, coarsening of rankings may cause a bias in the number of observed pairwise preferences. In particular, it may happen that some (pairs of) items are observed much more often than others. Therefore, in addition to the BTL problem as formalized above, we also consider the same problem with relative winning frequencies $\hat{p}_{i,j}$ instead of absolute frequencies $c_{i,j}$; we call this approach BTL(R).

\subsubsection{Least Squares/HodgeRank (LS)}

The HodgeRank algorithm \cite{hodgerank11} is based on a least squares approach. First, the probability matrix $\hat{P}$ is mapped to a matrix $X$ as follows:
\[
X_{i,j}=\begin{dcases*}
\log \left(\frac{\hat{p}_{i,j}}{\hat{p}_{j,i}} \right)  & \text{ if } $i \ne j$ and $\hat{p}_{j,i} \in (0,1)$,\\
0 & \text{ otherwise}
\end{dcases*}
\]
Then, $\hat{\pi} = \argsort (\theta^{*})$, where 
\[
\theta^{*} \in \arg \min_{\theta \in \mathbb{R}^{K} } \sum_{(i,j) \in \mathcal{E}} \Big ( (\theta_{j} - \theta_{i})  - X_{i,j} \Big )^{2},
\]
and $\mathcal{E}= \big \{ (i,j) \, | \, 1 \le i  < j \le K, \; X_{i,j} \ne 0 \big \}$.

\subsection{Voting Methods}

\subsubsection{Borda Count (Borda)}
Borda \cite{borda81} is a scoring rule that sorts items according to the sum of weighted preferences or ``votes'' in favor of each item: 
\[
\hat{\pi} = \argsort \{ s_{1}, \dots, s_{K} \} \enspace ,
\]
where $s_i =\sum_{i=1}^{K} \hat{p}_{i,j}$.

\subsubsection{Copeland (CP)}
Copeland \cite{copeland51} works in the same way as Borda, except that scores are derived from binary instead of weighted votes:
$$
s_i = \sum_{i=1}^{K} \mathbb{I} \left(\hat{p}_{i,j}>\frac{1}{2} \right) \enspace .
$$
%is another voting-based algorithm that counts the number of pairwise wins for each item $a_{i}$. Specifically speaking, it outputs the score $s(i)$ assigned to each item $a_{i}$ as $s(i)=\sum_{i=1}^{K} \mathbb{I} (p_{i,j}>\frac{1}{2})$.
%Then,
%\[
%\hat{\pi} = \argsort \{ s_{1}, \dots, s_{K} \}.
%\]

\subsection{Spectral Methods}

The idea of deriving a consensus ranking from the stationary distribution of a suitably constructed Markov chain has been thoroughly studied in the literature \cite{spectral49, spectral09, spectral98}. The corresponding Markov chain with transition probabilities $Q$ is defined by the pairwise preferences. Then, if $Q$ is an irreducible, aperiodic Markov chain, the stationary distribution $\bar{\pi}$ can be computed, and the predicted ranking is given by $\hat{\pi} = \argsort (\bar{\pi})$.

\subsubsection{Rank Centrality (RC)}
The rank centrality algorithm \cite{rc12} is based on the following transition probabilities:
\[
Q_{i,j}=\begin{dcases*}
\frac{1}{K} \; \hat{p}_{i,j}  & \text{ if } $i \ne j$ \\
1 - \frac{1}{K} \sum_{k \ne i} \hat{p}_{k,i} & \text{ if } $i=j$ 
\end{dcases*}
\]

\subsubsection{MC2 and MC3}
\citet{mc01} introduce four spectral ranking algorithms, two of which we consider for our study (namely MC2 and MC3), translated to the setting of pairwise preferences. For MC2, the transition probabilities are given as follows:
\[
Q_{i,j}=\begin{dcases*}
\dfrac{1}{\sum_{j=1}^{K} \hat{p}_{i,j}} \; \hat{p}_{j,i}  & \text{ if } $i \ne j$ \\
0 & \text{ if } $i=j$ 
\end{dcases*}
\]
The MC3 algorithm is based on the transition probabilities
\[
Q_{i,j}=\begin{dcases*}
\dfrac{1}{deg(a_{i})} \; \hat{p}_{i,j}  & \text{ if } $i \ne j$ \\
1 - \frac{1}{deg(a_{i})} \sum_{k \ne i} \hat{p}_{k,i} & \text{ if } $i=j$ 
\end{dcases*}
\]
where 
\[
deg(a_{i}) = \max \bigg ( \sum_{i=1}^{K} \mathbb{I} (\hat{p}_{i,j}>0) , \sum_{i=1}^{K} \mathbb{I} (\hat{p}_{j,i}>0) \bigg ).
\]

\subsection{Graph-based Methods}

\subsubsection{Feedback Arc Set (FAS)}
If the $c_{i,j}$ are interpreted as (weighted) preferences, the degree of inconsistency of ranking $a_i$ before $a_j$ is naturally quantified in terms of $c_{j,i}$. Starting from a formalization in terms of a graph whose nodes correspond to the items and whose edges are labeled with the weighted preferences, the weighted feedback arc set problem \cite{fas01, fas10} is to find the ranking that causes the lowest sum of penalties:
$$
\hat{\pi} = \argmin_{\pi \in \S_K} \sum_{(i,j): \,\pi(i)<\pi(j)} c_{j,i} 
$$
For the same reason as in the case of BTL, we also consider the FAS problem with edge weights given by relative winning frequencies $\hat{p}_{i,j}$ and binary preferences $\mathbb{I} (\hat p_{i,j} > 1/2)$ instead of absolute frequencies $c_{i,j}$; we call the former approach FAS(R) and the latter FAS(B).

\subsection{Pairwise Coupling}

A common approach to multi-class classification is the all-pairs decomposition, in which one binary classifier $h_{i,j}$ is trained for each pair of classes $a_i$ and $a_j$ \cite{fuer_rr02}. At prediction time, each classifier produces a prediction, which can be interpreted as a vote, or weighted vote $\hat{p}_{i,j}$ in case of a probabilistic classifier, in favor of item $a_i$ over $a_j$. The problem of combining these predictions into an overall prediction for the multi-class problem is also called \emph{pairwise coupling} \cite{hast_cb98}. 

To the best of our knowledge, pairwise coupling has not been used for rank aggregation so far. In fact, the original purpose of this technique is not to rank items but merely to identify a single winner. Nevertheless, since coupling methods are eventually based on scoring items, they can be generalized for the purpose of ranking in a straightforward way. Indeed, they have been used for that purpose in the context of label ranking \cite{mpub131}.

\subsubsection{Method by Hastie and Tibshirani (HT)}
\citet{hast_cb98} tackle the problem in the following way: Given relative frequencies $\hat{P}$, they suggest to find the probability vector $\vec{p}=(p_1,\ldots, p_K)$ that minimizes the (weighted) Kullback-Leibler (KL) distance 
\[
\ell(\vec{p}) = \sum_{i<j} n_{i,j} \bigg [ \hat{p}_{i,j} \log \frac{\hat{p}_{i,j}}{\mu_{i,j}} + 
(1-\hat{p}_{i,j}) \log \frac{1-\hat{p}_{i,j}}{ 1-\mu_{i,j} }
\bigg ]
\]
between $\hat{p}_{i,j}$ and $\mu_{i,j} = \frac{p_{i}}{ p_{i}+p_{j} }$, where $n_{i,j} = c_{i,j} + c_{j,i}$. To this end, the problem is formulated as a fixed point problem and solved using an iterative algorithm. 
Once $\vec{p}$ is obtained, the predicted ranking is determined by $\hat{\pi} = \argsort (\vec{p})$.

\subsubsection{Method by Price et al. (Price)}
\citet{price94} propose the following parameter estimation for each $i \in [K]$:
\[
\hat{\theta}_{i} = \dfrac{1}{ \big ( \sum_{j \ne i} \frac{1}{\hat{p}_{i,j} } \big ) - (K-2) }
\]
Then, the predicted ranking is given by $\hat{\pi} = \argsort (\hat{\theta})$.

\subsubsection{Method by Ting-Fan Wu et al. (WU1, WU2)}

\citet{wu04} propose two methods. The first one (WU1) is based on the Markov chain approach with transition matrix
\[
Q_{i,j}=\begin{dcases*}
\hat{p}_{i,j} / (K-1) & if $i \ne j$ \\
\sum_{s \ne i} \hat{p}_{i,s} / (K-1)  & if $i = j$
\end{dcases*} \enspace .
\]
Once the stationary distribution $\bar{\pi}$ of $Q$ is obtained, the predicted ranking is given by $\hat{\pi} = \argsort(\bar{\pi})$.
In their second approach (WU2), the following optimization problem is proposed:
\[
\min_{\theta} 2 \theta^{T} Q \theta \, ,
\]
where
\[
Q_{i,j}=\begin{dcases*}
- \hat{p}_{i,j} \hat{p}_{j,i} & if $i \ne j$ \\
\sum_{s \ne i} \hat{p}_{s,i}^{2}  & if $i = j$ 
\end{dcases*} \enspace .
\]
Once the optimization is solved, the predicted ranking is $\hat{\pi} = \argsort (\theta)$.

%\vspace{\sectionBefore}
\section{Practical Performance}
%\vspace{\sectionAfter}

\subsection{Synthetic Data}

To investigate the practical performance of the methods presented in the previous section, we first conducted controlled experiments with synthetic data, for which the ground truth $\pi^*$ is known. To this end, data in the form of pairwise observations was generated according to (\ref{eq:dgp}) with different distributions $\prob_\theta$ and $\prob_\lambda$, predictions $\hat{\pi}$ were produced and compared with $\pi^*$ in terms of the Kendall distance, i.e., the number of pairwise inversions:  
\[
\sum_{1 \le i < j \le K} \mathbb{I} \Big [ \text{sign}( \pi^{*}(i) - \pi^{*}(j) ) \ne \text{sign}( \hat{\pi}(i) - \hat{\pi}(j) ) \Big ].
\]
For each setting, specified by parameters $K, \theta, \lambda$, we are interested in the expected performance of a method as a function of the sample size $N$, which was approximated by averaging over 500 simulation runs.

\subsubsection{PL Distribution}

In a first series of experiments, synthetic data was produced for $K \in  \{ 3,4,5,7 \}$, $\prob_\theta$ the PL model with ground truth parameter $\theta \in \mathbb{R}^{K}_{+}$, and coarsening rankings by projecting to all possible pairs of ranks (i.e., using a degenerate distribution $\prob_\lambda$ with $\lambda_{i,j}=1$ for some $1 \le i < j \le K$). As a baseline, we also produced the performance of each method for the case of \emph{full} pairwise information about a ranking, i.e., adding all pairwise preferences to the data (instead of only $a_{\pi^{-1}(i)} \succ a_{\pi^{-1}(j)}$) that can be extracted from a ranking $\pi$.  

Due to space restrictions, the results are only shown in the supplementary material. The following observations can be made:
\begin{itemize}
	
	\item Comparing the results for learning from incomplete data with the baseline, it is obvious that coarsening comes with a loss of information and makes learning more difficult. 
	
	\item The difficulty of the learning task increases with decreasing distance $|i-j|$ between observed ranks and is most challenging for the case of observing neighboring ranks $(i,i+1)$.  
	
	\item All methods perform rather well, although FAS is visibly worse while BTL is a bit better than the others. On the one side, this could be explained by the fact that the BTL model is consistent with the PL model, as it corresponds to the marginals of $\prob_\theta$. On the other side, like all other methods, BTL is agnostic of the coarsening $\prob_\lambda$; from this point of view, the strong performance is indeed a bit surprising. As a side remark, BTL(R) does not improve on BTL.

	\item While FAS is on a par with FAS(R), FAS(B) tends to do slightly better, especially for a larger number of items. However, as already said, FAS performs generally worse than the others.
\end{itemize}
In another set of experiments, we examined the \emph{averaged} performance of methods over all coarsening positions. The results are again shown in the supplementary material. It is noticeable that, as the number of items increases, the performance of the FAS-based approaches decreases. Moreover, as pointed out earlier, the BTL performs moderately better than other approaches.

\subsubsection{Mallows Distribution}

In a second series of experiments, we replaced the PL distribution with another well-known distribution on rankings, namely the Mallows distribution. Thus, data is now generated with $\prob_\theta$ in (\ref{eq:dgp}) given by Mallows instead of PL. This experiment serves as a kind of sensitivity analysis, especially for those methods that (explicitly or implicitly) rely on the assumption of PL. 

The Mallows model \cite{mallows57} is parametrized by a reference ranking $\pi^*$ (which is the mode) and a dispersion parameter $\phi$, i.e., $\theta=(\pi^*, \phi)$:
\[
\prob_{\pi^* , \phi} (\pi) = \dfrac{1}{Z(\phi)} \exp \big(
-\phi  D(\pi, \pi^*) \big),
\]
where $D(\pi, \pi^*)$ is the Kendall distance and $Z(\phi)$ a normalization constant. 

The results and observations for this series of experiments are quite similar to those for the PL model. What is notable, however, is a visible drop in performance for BTL, whereas Copeland ranking now performs much better than all other algorithms. Furthermore, FAS(B) is significantly better than FAS and FAS(R). These results might be explained by the ordinal nature of the Mallows model, which is arguably better fit by methods based on binary comparisons than by score-based approaches.

\begin{figure}
	\begin{center}
		\vspace*{0mm}
		\includegraphics[scale=0.3]{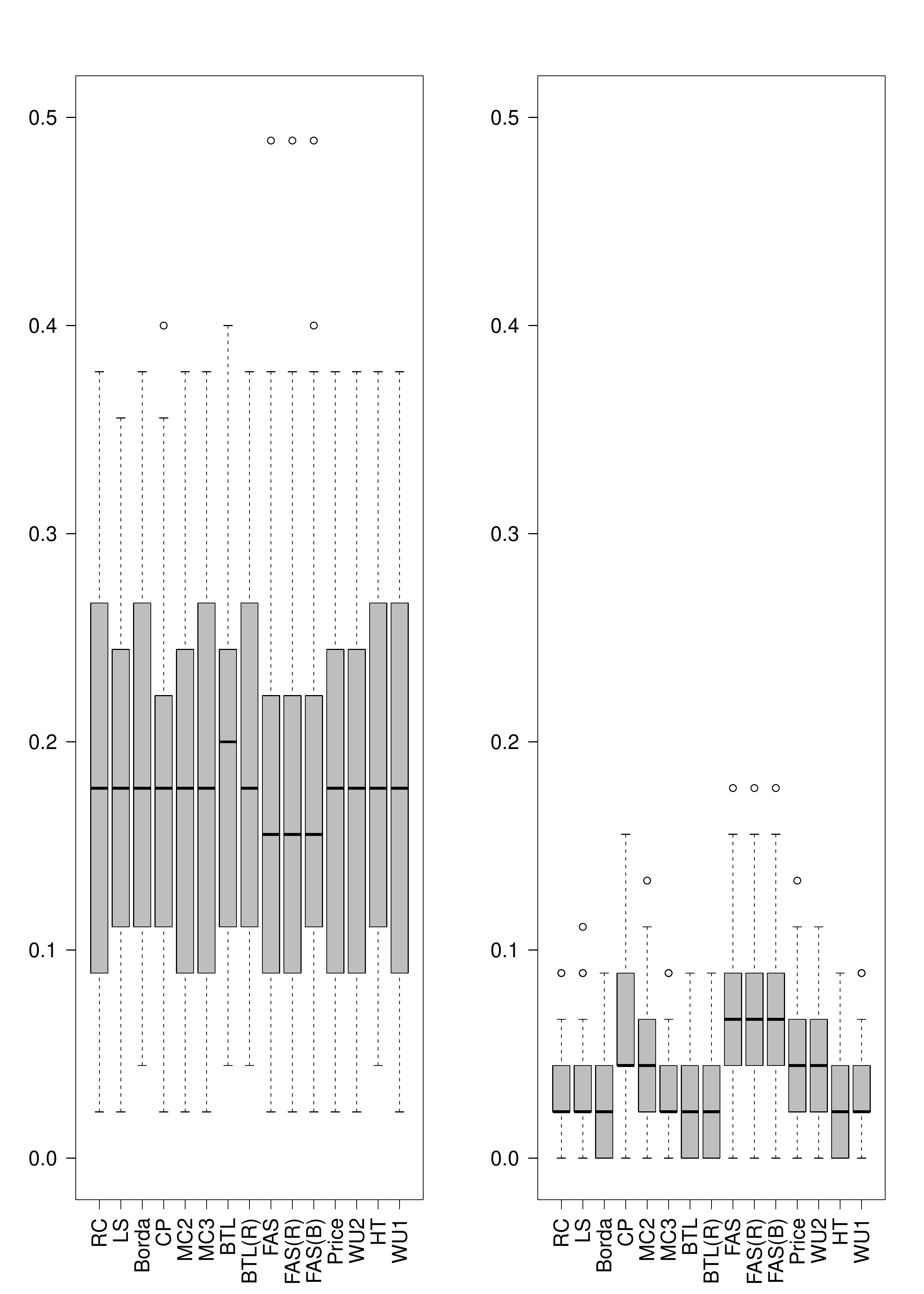}
		\caption{Results for the sushi data in terms of normalized Kendall distance (one boxblot per method), experiment (a) on the left and (b) on the right.}
		\label{fig:sushi}
	\end{center}
\end{figure}

\subsection{Real Data}

To compare the methods on a real world data set, we used the Sushi data \cite{kamishima03} that contains the preferences (full rankings) of 5000 people over 10 types of sushi. 

For this data set, there is no obvious ground truth $\pi^*$. Therefore, we define a separate target for each data set individually, which is the ranking produced by that method on the full set of pairwise comparisons that can be extracted from the training data. The distance between this ranking and the one predicted for coarsened data can essentially be seen as a measure of the loss of information caused by coarsening. We conduct this experiment for (a) degenerate coarsening where $\lambda_{i,j} = 1$ for some $(i,j)$ and (b) random coarsening where $\lambda _{i,j} = 2/(K^2-K)$ for all $(i,j)$. Note that, while the information content is the same in both cases (one pairwise comparison per customer), random coarsening does not introduce any bias, as opposed to degenerate coarsening. 

The results (distributions for 100 repetitions) are shown in Figure \ref{fig:sushi}. Again, all methods are more or less on a par. However, as expected, random coarsening does indeed lead to better estimates, again suggesting that ``real'' coarsening indeed makes learning harder.

%\vspace{\sectionBefore}
\section{Consistency}
%\vspace{\sectionAfter}

Recall the (specific) setting we introduced in Section \ref{sec:specific}, in which full rankings are generated according to a PL model with parameter $\theta = (\theta_1, \ldots , \theta_K)$, and the coarsening corresponds to a projection to a random pair of ranks; for technical reasons, we subsequently assume $\theta_i \neq \theta_j$ for $i \neq j$. Also recall that we denote by $p_{i,j}$ the probability of a ranking $\pi$ in which $a_i$ precedes $a_j$, and by $q_{i,j}$ the probability to observe the preference $a_i \succ a_j$ after coarsening.  According to PL, we have $p_{i,j} = \theta_i/(\theta_i + \theta_j)$, and the ground truth ranking $\pi^*$ is such that 
$$
(\pi(i) < \pi(j)) \, \Leftrightarrow \, (\theta_i > \theta_j)  \, \Leftrightarrow \, (p_{i,j} > 1/2)
$$
Finally, we denote by $\hat{p}_{i,j}$ the estimate of the pairwise preference (probability) $p_{i,j}$. If not stated differently, we always assume estimates to be given by relative frequencies, i.e., $\hat{p}_{i,j} = w_{i,j}/(w_{i,j} + w_{j,i})$, with $w_{i,j}$ the observed number of preferences $a_i \succ a_j$.

\textbf{Definition 1:}
Let $\hat{\pi}_N$ denote the ranking produced as a prediction by a ranking method on the basis of $N$ observed (pairwise) preferences. The method is consistent if $\prob(\hat{\pi}_N = \pi^*) \rightarrow 1$ for $N \rightarrow \infty$.

The proofs of the following results are given in the supplementary material.

\textbf{Lemma 2:}
\label{le:swap}
Let us consider a probability measure $\p_\theta$  over $\S_K.$ Consider $q_{i,j}=\sum_{\pi \in E(a_i\succ a_j)}\p_\theta(\pi) \lambda_{\pi(i),\pi(j)}, \ \forall\, i\neq j$. (The model (\ref{eq:pcp}) with $\p_\theta(\pi)$ not necessarily PL).
If $\p_\theta(\pi)\geq \p_\theta(\pi_{i,j})$ for all $\pi\in E(a_i\succ a_j)$, then $q_{i,j}>q_{j,i}$.

\textbf{Lemma 3:}
\label{le:qp}
Assume the model (\ref{eq:pcp}), $\theta_i \neq \theta_j$ for $i \neq j$, and $\theta_i > 0$ for all $i \in [K]$. The coarsening (\ref{eq:pc}) is order-preserving for PL in the sense that $p_{i,j} > 1/2$ if and only if $q_{i,j}' > 1/2$, where $q_{i,j}' = q_{i,j}/(q_{i,j}+q_{j,i})$.

The last result is indeed remarkable: Although coarsening will bias the pairwise probabilities $p_{i,j}$, the ``binary'' preferences will be preserved in the sense that $\sign(p_{i,j} - 1/2) = \sign(q_{i,j}' - 1/2)$. Indeed, the result heavily exploits properties of the PL distribution and does not hold in general. For example, consider a distribution $\prob$ on $\S_3$ such that $\prob([1,2,3])=0.8$, $\prob([3,1,2]) = \prob([3,2,1])=0.1$. Then, with a coarsening (\ref{eq:pc}) such that $\lambda_{2,3}=1$, we have $p_{1,2} = p_{1,3} = 0.8$, but $q_{1,2}' = q_{1,3}' = 0$.

\textbf{Lemma 4:}
\label{le:slln}
Assume the model (\ref{eq:pcp}), $\theta_i \neq \theta_j$ for $i \neq j$, and $\theta_i > 0$ for all $i \in [K]$. Let us take an arbitrarily small $\epsilon^*>0.$ There exists $N_0\in\mathbb{N}$ such that  $\theta_i > \theta_j$  if and only if $\hat{p}_{i,j} > 1/2$ for all $i,j \in [K]$, with probability at least $1-\epsilon^*$,  after having observed at least $N_0$ preferences.

\textbf{Theorem 5:}
Copeland ranking is consistent.

\textbf{Theorem 6:}
FAS, FAS(R), and FAS(B) are consistent.

Our experimental results so far suggest that consistency does not only hold for Copeland and FAS, but also for most other methods (including BTL), and hence that rank-dependent coarsening is indeed somehow ``good-natured''. Anyway, for these cases, the proofs are still pending. % which are often highly non-trivial.

%\vspace{\sectionBefore}
\section{Summary and Conclusion}
%\vspace{\sectionAfter}

In this paper, we addressed the problem of learning from incomplete ranking data and advocated an explicit consideration of the process of turning a full ranking into an incomplete one---a step that we referred to as ``coarsening''. To this end, we proposed a suitable probabilistic model and introduced the property of rank-dependent coarsening, which can be seen as orthogonal to standard marginalization: while the latter projects a ranking to a subset of items, the former projects to a subset of ranks. 

First experimental and theoretical results suggest that agnostic learning can be successful under rank-dependent coarsening: even if ignorance of the coarsening may lead to biased parameter estimates, the ranking task itself can still be solved properly. This applies at least to the specific setting that we considered, namely rank aggregation based on pairwise preferences, with Plackett-Luce (or Mallows) as an underlying distribution.

Needless to say, this paper is only a first step. Many questions are still open, for example regarding the consistency of ranking methods, not only for the specific setting considered here but even more so for generalizations thereof. In addition to theoretical problems of that kind, we are also interested in practical applications such as ``crowdordering'' \cite{mats_c14,chen_pr13}, in which coarsening could play an important role.

%\bibliography{example_paper}
\bibliography{lit}
\bibliographystyle{icml2017}

\end{document}